\documentclass[lettersize,journal]{IEEEtran}
\usepackage{amsmath,amsfonts}
\usepackage{algorithmic}
\usepackage{algorithm}
\usepackage{array}
\usepackage[caption=false,font=normalsize,labelfont=sf,textfont=sf]{subfig}
\usepackage{textcomp}
\usepackage{stfloats}
\usepackage{url}
\usepackage{verbatim}
\usepackage{graphicx}
\usepackage{cite}
\usepackage[encapsulated]{CJK}
\usepackage{multirow}
\usepackage{booktabs}
\usepackage{color,xcolor}
\usepackage{multirow}

\begin{document}

\title{LipFormer: Learning to Lipread Unseen Speakers based on Visual-Landmark Transformers}

\author{Feng Xue, Yu Li, Deyin Liu, Yincen Xie, Lin Wu,~\IEEEmembership{Senior Member, ~IEEE,} Richang Hong*, ~\IEEEmembership{Senior Member,~IEEE}
\thanks{F. Xue, Y. Li, Y. Xie, L. Wu, R. Hong are Hefei University of Technology, China. D. Liu is with Anhui University, China}
\thanks{Corresponding author: Richang Hong}
}



\maketitle

\begin{abstract}
Lipreading refers to understanding and further translating the speech of a speaker in the video into natural language. State-of-the-art lipreading methods excel in interpreting overlap speakers, i.e., speakers appear in both training and inference sets. However, generalizing these methods to \textit{unseen} speakers incurs catastrophic performance degradation due to the limited number of speakers in training bank and the evident visual variations caused by the shape/color of lips for different speakers. Therefore, merely depending on the visible changes of lips tends to cause model overfitting. To address this problem, we propose to use multi-modal features across visual and landmarks, which can describe the lip motion irrespective to the speaker identities. Then, we develop a sentence-level lipreading framework based on visual-landmark transformers, namely \textit{LipFormer}. Specifically, LipFormer consists of a lip motion stream, a facial landmark stream, and a cross-modal fusion. The embeddings from the two streams are produced by self-attention, which are fed to the cross-attention module to achieve the alignment between visuals and landmarks. Finally, the resulting fused features can be decoded to output texts by a cascade seq2seq model. Experiments demonstrate that our method can effectively enhance the model generalization to unseen speakers.
\end{abstract}

\begin{IEEEkeywords}
Lipreading, Landmarks, Transformer, Lip motion.
\end{IEEEkeywords}

\section{Introduction}\label{sec:introduction}

\IEEEPARstart{L}{ipreading} is the inference on the speech of a speaker from a video clip, which could be presented with/without audial signals \cite{2019Hearing,zhang2019understanding,hilder2009comparison}. Lipreading offers an effective way to infer text, alternative to speech recognition, which renders implausible in disturbing circumstances, e.g, unknown speakers in the wild. Besides, lipreading shows enormous values to real-world applications, such as silent-movie processing and silent conversations \cite{KIM2004295,2004Lip,2020Discriminative}.

\begin{figure}[!t]
\centering
\includegraphics{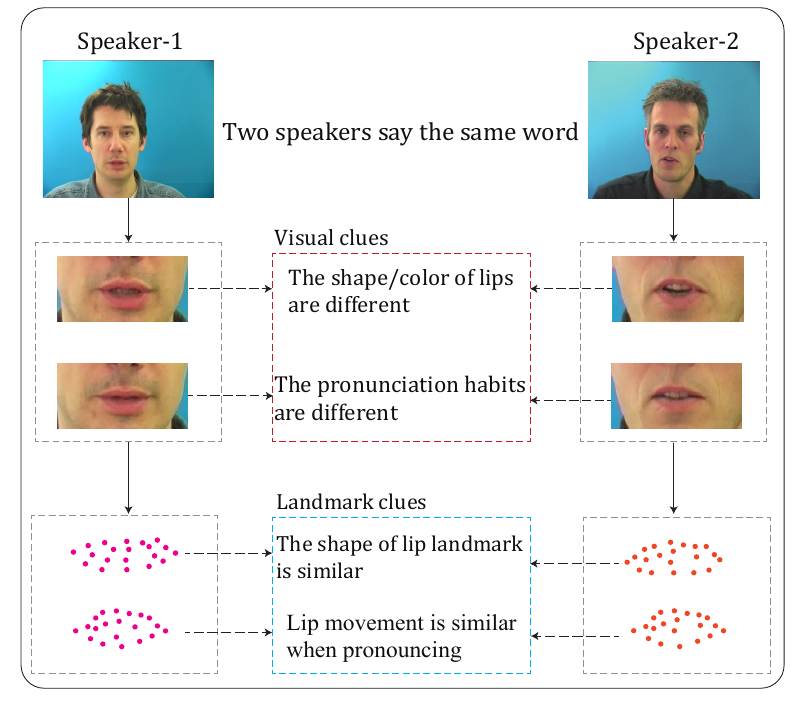}
\caption{ The visual variations of lip motion can easily cause the overfitting of a lipreading model by associating spurious correlation between motion and texts. In this paper, we propose to use landmarks to calibrate the cross-modal association.}
\label{fig:motivation}
\end{figure}

Benefited by deep learning, lipreading has also witnessed its remarkable progression, which has demonstrated its trend to even surpass experienced subject experts. Early efforts are made to perform lipreading only at word-level \cite{petridis2018end,stafylakis2017combining}. However, such lipreading method only corresponds to one word at each time. Compared to word-level lipreading, sentence-level lipreading \cite{zhang2021efficient,zhao2020mutual,zhao2019cascade,xu2018lcanet,liu2020fastlr} is more accurate in sentence prediction by predicting the texts depending on the contextual priors. For example, Assael \textit{et.al}\cite{assael2016lipnet} proposed LipNet, which combines VGG\cite{2014Return}, LSTM\cite{2014Empirical}, and CTC\cite{0Connectionist}, and thus achieved an accuracy of 95.2 on the GRID dataset\cite{cooke2006audio}. In\cite{huang2021callip}, the authors developed an approach based on attribute learning and contrast learning, which greatly improved the performance of lipreading. However, the majority of current lipreading models are only trained and tested on publicly available datasets, which are limited in their training sample size and number of speakers. Moreover, the performance improvement of these methods are incremental to unseen speakers as they are mainly developed in the case of overlap speakers. (i.e., the test speaker has ever appeared in the training set). We hypothesize that these methods describe the lip motion using visual clues, however, if the model is only trained with visual lip motion, it will cause the overfitting due to the visual variations caused by the shape/color of lips and pronunciation habit of a particular speaker. As a consequence, this hampers the generalization ability of the model. For example, as shown in Figure \ref{fig:motivation}, being overfitted to the visual variations e.g., the lip shape, the model translates into different texts even when two speakers say the same word. Therefore, developing a lipreading method simply using lip motion may slump the translation accuracy especially in unseen speakers. In real-life applications, a lipreading system is often required to make lip-to-text predictions for new faces, which may not be observed in the training bank. Also, learning a model with good generalization ability to unseen speaks is paramount to downstream applications.

\begin{figure*}[hbt]
\centering
\includegraphics{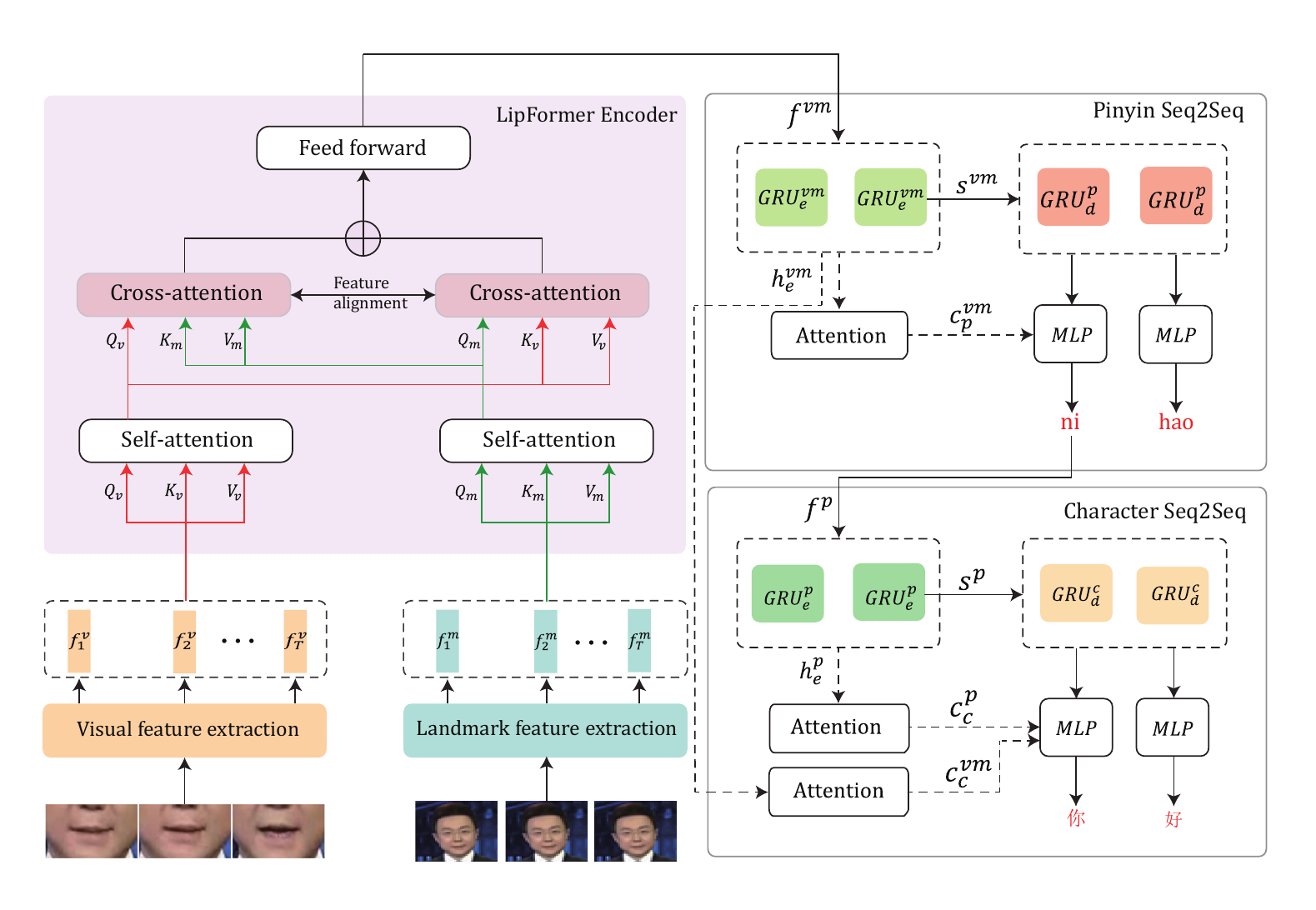}
\caption{ The proposed LipFormer is an end-to-end two-stream architecture built upon a visual branch and a landmark branch. The input to the visual branch is a sequence of lip images. The input to the landmark branch is a 340-dimensional vector extracted from speaker frame. The embeddings from the two streams are forwarded to a cross-attention module to achieve the alignment, which allows for the cross-modal fusion. The resulting fused features can be decoded to output texts by a cascaded seq2seq model. Definitions of notations can be seen in Table \ref{tab:sym}.}
\label{fig:model}
\end{figure*}

Given limited access to the number of samples and speaker identities, we aim to enhance the model in lipreading by calibrating the association between motion and texts via \textit{corrective alignment}, such that the model can generalize to unseen speakers. In this paper, we explore to use landmarks as a corrective offset to achieve the true underlying association between lip motion and sentences for lipreading. In \cite{zhang2018unsupervised}, the author proposed to learn object structural representations by using landmarks as a complementary feature to the pretrained deep representation in recognizing visual attributes. In \cite{haghpanah2022real}, the authors proposed to extract facial geometric features by using landmarks, and both geometric and texture-based features can be used to improve the accuracy of facial expression recognition. Such a method can also help the model generalize to new faces, which are out of the training set. Inspired by those approaches, in this paper, we propose to introduce landmark features independent to the visual appearance of lips, and thus can eliminate the visual variations between speakers. In fact, landmarks encode positional priors of the speaker's face and lips (i.e., lip and facial landmarks). The motion trajectory encoded by those landmarks effectively describes lip motion irrespective to speakers. And the learned landmark embedding is less influenced by the visual variations. It shows potential to better the generalization of a lipreading model to the unseen speakers in inference.

In this paper, we aim to improve the generalization of a lipreading model in recognizing unseen speakers. To achieve this goal, we propose a sentence-level lipreading framework based on visual-landmark transformers, dubbed LipFormer. Specifically, we describe the lip motion with features from two modalities: visual features and landmarks. The visual features are extracted from lip regions for each speaker at frame level, and then self-attended to be discriminative. However, encoding the lip motions only using visual information can easily lead to over-fitting, owing to the strong bias towards visual variations caused by the shape/color of a speaker's lips. To this end, we propose to use landmarks extracted from face and lips as motion trajectory to eliminate such variations. Then, we employ the cross-attention to align and fuse such cross-modal features. The cross-attention of transformer can effectively learn the correspondence between the visual and landmark embeddings, so as to improve the joint representations for cross-modal fusion. Finally, we employ a cascaded sequence-to-sequence (seq2seq) to decode the fused features and generate the texts.

The main contributions of this paper are summarised below:

1) We propose a sentence-level lipreading framework based on visual-landmark transformers, which introduces corrective landmarks to minimize the biased visual variations, making the model generalize to unseen speakers. 

2) The proposed model uses cross-modal features to describe lip motion, and a cross-attention is adopted to achieve the alignment between visuals and landmarks, which promotes the fusion of heterogeneous features and further improves the generalization of the model. 

3) Extensive experiments are conducted on benchmark datasets to demonstrate the effectiveness of the proposed method in interpreting unseen speakers and a SOTA performance is achieved.

\begin{table}[hbt]
\fontsize{8}{8}\selectfont
\caption{Notations and Definitions.}
\label{tab:sym}
    \begin{tabular}{ll}
        \toprule
      \textbf{Symbol} & \textbf{Definition}\\
        \midrule
        $ f^{v}_{T}$, $ f^{m}_{T}$& visual feature, landmark feature\\ 
        $ f^{vm}$& visual-landmark embedding sequence \\
        $ f^{p}$&pinyin embedding sequence \\
        $ GRU_{e}$ &The subscript e indicates the encoder\\
        $ GRU_{d}$&The subscript d indicates the decoder\\
        $ GRU^{vm}_{e}$& GRU unit in visual-landmark encoder  \\
        $ GRU^{p}_{e}$, $ GRU^{p}_{d}$& GRU unit in pinyin encoder and pinyin decoder  \\
        $ GRU^{c}_{d}$& GRU unit in character decoder  \\
        $ h^{vm}_{e}$, $ h^{p}_{e}$&visual-landmark encoder output, pinyin encoder output   \\
        $ c^{vm}_{p}$, $ c^{p}_{c}$, $ c^{vm}_{c}$& context vector calculated by the attention  \\
    
        \bottomrule
    \end{tabular}
\end{table}

\section{Related work}
 In this section, we briefly review the lipreading methods, which can be categorized into two lines: conventional methods and deep learning-based lipreading methods.

\subsection{Traditional lipreading methods}
1) Pixel-based methods. They assumes that all pixels in the lip region contain vision-related information, and uses the pixel value of the lip region as the original features. The features are reduced in different ways to obtain expressive features. For example, Potamianos et al.\cite{2012On} proposed HiLDA, which is widely used as a visual feature extractor in speech recognition tasks. Lucey et al.\cite{2008Patch} further considered local features based on this, the author extracted local features of image patch, fusing global features with local features to further improve recognition accuracy. Tim et al.\cite{sheerman2011cultural} normalize and concatenate the AAM features of consecutive frames to extract spatio-temporal features by linear transformation. 2) Shape-based methods. They extract features based on the shape of the lip region (lips, chin, etc.). For example, Papcun et al.\cite{1992Inferring} used articulatory features (AFs) for lipreading, but since this kind of features is too simple to distinguish similar word, it is generally applied to small-scale recognition tasks. Chan\cite{2001Hmm} combined geometric features with PCA features of the lip as visual features. Luettin et al. \cite{1997Speechreading} applied the ASM model to lipreading, generating features from the coordinates of several key points. However, the shape-based model assumes that most of the information related to visual is on the contour represented by feature points, which inevitably leads to information loss.

Benefited by deep learning, lipreading has also witnessed its remarkable progression. Compared with traditional methods, deep learning-based methods has powerful feature learning capability. The deep learning method avoids the complex hand-crafted feature extraction process, and the performance of its model can be further enhanced with large-scale data.

\subsection{Deep learning-based lipreading methods}
Lipreading can be carried out at word-level and sentence-level lipreading. Early efforts are made to perform lipreading only at word-level, such lipreading video only corresponds to one word and had a small range of applications. For example, Chung \cite{10.1007/978-3-319-54184-6_6} designed two CNN structures, i.e., early fusion and multiple towers, which can be combined for an all-once word-level translation from a sequence of lip motion. Petridis et al \cite{petridis2018end} proposed an end-to-end audio-visual model based on residual networks and BiGRU, which simultaneously learns to extract features directly from the image and audio. Stafylakis et al\cite{stafylakis2017combining} combined 3D-CNN and 2D-CNN to extract visual features and obtained higher accuracy on the LRW dataset.

In this paper, we focus on sentence-level lipreading. Compared to word-level lipreading, sentence-level lipreading is more accurate, since it predicts the texts depending on the contextual priors. The LipNet \cite{assael2016lipnet} is the first end-to-end sentence-level lipreading, which consists of 3DCNN, BiGRU, and CTC. LipNet achieves 95.2\begin{math} \%  \end{math} accuracy on the GRID dataset. In \cite{2018END,2017Improving,2018Investigations}, the model structure is similar to LipNet. However, CTC loss has conditional independence, i.e., each output unit is individually predicting the probability of a label. Therefore, CTC loss will focus on local information of adjacent frames, which is not suitable for predicting labels that require contextual information to discriminate. Considering the problems incurred by CTC loss, Xu et al.\cite{xu2018lcanet} proposed LCANet, which stacks two layers of Highway networks in 3DCNN. This can highly improve the quality of the extracted features. They essentially used attention mechanism to overcome the shortcomings of conditional independence assumption in CTC. The following work \cite{huang2021callip} improves the performance of the lipreading model by introducing attribute learning and contrast learning into the sentence-level lipreading pipeline.

Other lipreading methods are based on the seq2seq model. The most representative model is WAS\cite{son2017lip}, which uses a 5-layer 2DCNN and LSTM to extract visual features and auditory features. These features are fed to the seq2seq module to generate texts. Zhao \cite{zhao2019cascade} proposed CSSMCM, which combines factors such as pinyin and tones to help predict Chinese characters based on visual information. In \cite{2019Hearing}, the authors proposed a knowledge distillation method that uses a speech recognition pre-trained model as a teacher model to optimize a lipreading model as a student model, improving the accuracy of lipreading.

Due to the excellent performance of Transformer, Zhou et al.\cite{zhou2018syllable} apply transformer to speech recognition. Ma et al.\cite{2020A} apply the transformer to lipreading, and thus proposed the CTCH-LipNet, which first used 3DCNN to extract visual features, and then a cascaded architecture that consists of two transformers to predict pinyin and Chinese characters. Ma et al.\cite{2022Visual} use both video and audio as input, and the transformer decodes the features into texts.

However, most existing lipreading methods use overlap speakers by default in experimental evaluation. This is prone to be overfitting to overlap speakers in the training set, while perform poorly to the unseen speakers. In this paper, we propose a sentence-level lipreading framework based on visual-landmark transformers that generalize the model to unseen speakers, thus solving the lipreading problem of unseen speakers.

\section{The Method}\label{sec:method}

The architecture of our proposed LipFormer is shown in Fig \ref{fig:model}, which consists of four modules: 1) a visual stream that extracts visual features of lip regions; 2) a landmark stream that encodes the trajectory of lip/facial movement of a speaker across the sequence; 3) a cross-modal fusion that learns the alignment between visuals and landmarks; and 4) a cascades seq2seq model for mapping fused features to texts.

\subsection{Visual Embedding}
 For each video clip, we first extracted the face image in each frame by using the DLib face detector, and then apply an affine transformation to each face image to obtain the mouth-centered cropping with 160 × 80 pixel as the lip region. For a video clip with $T$ frames, we can have a lip region sequence $\{T_i\}$, where $I_{i}$ is the frame of the $i$th step ($i=1,\ldots,T$). We first learn the per-frame feature embedding by applying the 3D convolution \cite{3D-PersonVLAD}, followed by a ReLu layer and a max-pooling layer. During training, dropout regularization is used along with the 3DCNN to alleviate the saturation problem. 
 
\begin{figure}[!t]
\centering
\includegraphics[width=3 in]{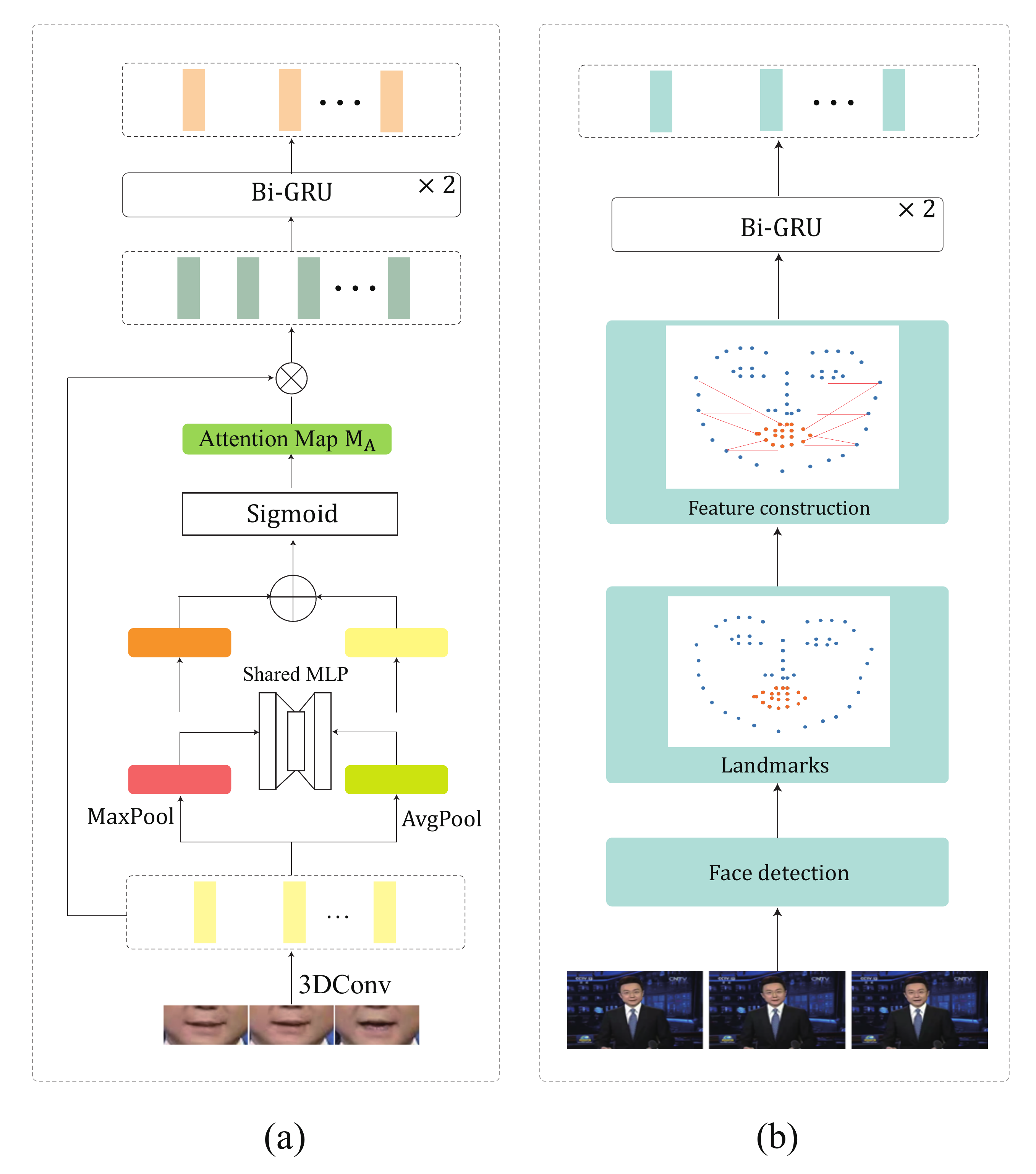}
\caption{ The two-stream architecture with a visual branch (a) and a landmark branch (b). Given the input as a sequence of lip regions, we use 3DCNN combined with channel attention to extract the features, followed by Bi-GRU to encode temporal orders. The input to the landmark branch is the landmark embedding: the difference of the angle matrix between adjacent frames. The angle matrix is calculated between 20 landmarks of the lip and 17 landmarks of the face contour for each frame and encoded as a 340-dimensional vector.}
\label{fig:two-branch}
\end{figure}

The visual features extracted by 3DCNN contain a lot of irrelevant information, such as the shape of lip, pronunciation habits of different speakers, etc. That information largely affects the accuracy of text generated by the decoder. In this paper, we combine 3DCNN with channel attention. The channel attention mechanism can learns the weights of each channel and improves the performance of useful visual features by suppressing irrelevant features. The obtained feature vector for each frame is denoted by $f^{v}_{i}$.  

To aggregate the spatial grids, two different spatial context descriptors are generated from the input feature map by using average pooling and maximum pooling, and these two outputs are fed into a shared network MLP to generate the channel attention map. Multiply the attention map and the input features to focus on important features. The structure of the visual branch is shown in Fig.3 (a).

Limited by the size of convolution kernel, CNN can only extract short-term spatio-temporal visual features. The bi-directional GRU \cite{WU-Co-attention,Chen-MM-2022} is applied to extract the long-term features:
 \begin{equation}
[s^{v}_{i},o^{v}_{i}]=GRU(f^{v}_{i},s^{v}_{i-1}),
 \end{equation}
where $s^{v}_{i}$, $o^{v}_{i}$ are the hidden vector and output vector of the GRU, respectively.

\subsection{Landmark Embeddings}
Merely depending on the visual appearance will lead to inferior performance for lipreading. A primary reason is that lip motion has a significant visual variation caused by the very different shape and color of the lip or a particular pronunciation habit of a speaker. To eliminate the visual variations, we propose taking the landmarks as another feature embedding. The landmark embedding will be less influenced by the lip appearance and has better generalization ability to the unseen speakers during the training. The distribution of the 68 facial landmarks is shown in Fig. 3(b), among which there are 20 landmarks for lip, 17 landmarks for facial contour, and 31 landmarks for eyes, eyebrows, and noses. The structure of the landmark branch is shown in Fig. \ref{fig:two-branch} (b).

More specifically, we utilize both lip landmarks and facial contour landmarks to construct the landmark embedding for lipreading. The change of facial contour position is found to be the most obvious with the lip motion. The motion trajectory between facial landmarks effectively describes lip motion. The features are constructed as follows: 1) We first calculate the angle between 20 landmarks of the lip and 17 landmarks of the face contour for each frame, i.e. the cosine, to obtain the angle matrix of (1,340); 2) We then compute the difference angle matrix for two adjacent frames to represent the motion change of landmarks. 3) For the T-frame, the (B,T,340) angle matrix is obtained as the input to the landmark branch. The matrix difference is the per-frame landmark embedding, denoted by  $ f^{m}_{i}$ for the $i$th frame. Given $ \left \{ f^{m}_{1}, f^{m}_{i},...f^{m}_{T}\right \} $ of all the $T$ frames, we apply a bi-directional GRU to extract the long-term features so as to obtain the output feature sequence $ \left \{ s^{m}_{1}, s^{m}_{i},...s^{m}_{T}\right \} $.

\subsection{Cross-Modal Fusion via Transformer}


The embeddings from the two streams are fed to the transformer. We use the transformer-encoder to achieve the cross-modal fusion. The encoder consists of an encoder layer, which is composed of three parts: a self-attention module, a cross-attention module, and a feed-forward network. The embeddings from the two streams are produced by self-attention, which are fed to the cross-attention module to achieve the alignment between visuals and landmarks. Three matrices of query $Q$, key $K$, and value $V$ as input to the self-attention, which are generated from the input sequence $z$. Self-attention module extracting global information to establish global long-term dependency of lip motion. Cross-attention takes the Q of the current modality (e.g., video) and the K/V obtained in the opposite modality (e.g., landmark) as input, for each visual feature embedding, different weights are assigned to each landmark feature embedding by cross-attention, and the matching of weights can achieve visual-landmark embedding alignment to achieve cross-modal feature fusion. The output of the attention module is:

\begin{equation}
 Attention=\textbf{softmax}\left (\frac{Q^i{(K^i)^T}}{\sqrt{d} }  \right ) \cdot V^i.
 \end{equation}
 where $d$ is the length of the embedding vector.
 
 The encoder layer is implemented as a feed-forward network, which contains two fully-connected layers with a ReLU non-linearity:
  \begin{equation}
 FFN(x)=FC(ReLU(FC(x))).
 \end{equation}

\subsection{Text Generation}
With the fused features as input, we feed them to a cascade sequence-to-sequence model. The seq2seq model is with an encoder-decoder structure, where both the encoder and decoder are LSTM models (sometimes GRU models). Encoder-Decoder model can predict arbitrary sequence correspondence. In the Chinese dataset, there are fewer pinyin categories, making it easier to predict the pinyin. For this reason, we choose pinyin as a middle layer when predicting Chinese characters and use a cascaded seq2seq model to decode text.

\subsection{Pinyin Prediction} 
With the fused features $ f^{vm} $ as input, we feed them to a Pinyin seq2seq model to decode the $ f^{vm} $ into pinyin. We refer to the encoder and decoder as visual-landmark encoder and pinyin decoder, in which the encoder processes the visual-landmark feature sequence, and the decoder predicts the pinyin sequence. The GRU of encoder calculating the hidden layer vector by inputting $ f^{vm} $ as:
\begin{equation}
(h^{vm}_{e})_{i}=GRU^{vm}_{e}((h^{vm}_{e})_{i-1},e^{vm}_{i}). 
\end{equation}

The decoder progressively computes the hidden layer vector $\left ( h^{p}_{d} \right ) _{i}$ based on the predicted result $p_i$ of the previous time step and the hidden state $\left ( h^{p}_{d} \right ) _{i-1}$of the previous time step:
\begin{equation}
(h^{p}_{d})_{i}=GRU^{p}_{d}((h^{p}_{d})_{i-1},E(p_{i})),
\end{equation}
where the embedding function $E(\cdot)$ maps $ p_{i} $ to its vector space. To further utilize the information contained in the input vector, we apply the attention module to compute the weights of all hidden layer vectors in the encoder to generate a context vector, that assists the decoder predicting the pinyin. The weights are calculated as: 
\begin{equation}
att=\textbf{softmax}(FC(tanh(FC(Concat((h^{p}_{d})_{i}),h^{p}_{e})))).
\end{equation}
Finally, the probability distribution of the current pinyin is calculated by splicing the GRU output and the context vector as:
\begin{equation}
P(p_{i})= \textbf{softmax}(FC((h^{p}_{d})_{i},h^{p}_{e}\cdot att.
\end{equation}

\subsection{Character Prediction} Similarly, With the pinyin sequence as input, we feed them to a character seq2seq model to translating pinyin to character, the computation process is similar to that of predicting pinyin sequences. First, the pinyin sequence is input to the pinyin encoder, and the GRU of the pinyin encoder calculating the hidden layer vector as:
\begin{equation}
(h^{p}_{e})_{i}=GRU^{p}_{e}((h^{p}_{e})_{i-1},e^{p}_{i}). 
\end{equation}
In addition, a dual attention mechanism is employed in the process of predicting characters in order to take both visual-landmark and pinyin information. The character decoder calculates the context vector based on the visual-landmark encoder output and the pinyin encoder output, and then predicts the character:
\begin{equation}
(h^{c}_{d})_{i}=GRU^{c}_{d}((h^{c}_{d})_{i-1},E(c_{i})). 
\end{equation}
\begin{equation}
c^{vm}_{i}=h^{vm}_{e}\cdot att((h^{c}_{d})_{i},h^{vm}_{e})). 
\end{equation}
\begin{equation}
c^{p}_{i}=h^{p}_{e}\cdot att((h^{c}_{d})_{i},h^{p}_{e})). 
\end{equation}
\begin{equation}
P(c_{i})= \textbf{softmax}(FC((h^{c}_{d})_{i},c^{vm}_{i},c^{p}_{i}).
\end{equation}

\subsection{Loss Function}
To improve the prediction accuracy, the model first predicts pinyin and then translating pinyin into Chinese characters. We jointly optimize the loss function of these two processes. The loss function defined as:
\begin{equation}
L=L_{p}+L_{c}, 
\end{equation}
where \begin{math} L_{p}=- {\textstyle \sum_{n=1}^{N}} log P(p_n|x,p_1,p_2,...p_{n-1})
\end{math}, \begin{math} L_{c}=- {\textstyle \sum_{n=1}^{N}} log P(c_n|x,c_1,c_2,...c_{n-1})
\end{math}
\section{Experiments}
In this section, we validate the proposed method by conducting extensive experiments on two benchmark datasets: CMLR \cite{zhao2019cascade} and GRID \cite{cooke2006audio}.

\begin{figure}[!t]
\centering
\includegraphics[width=3in]{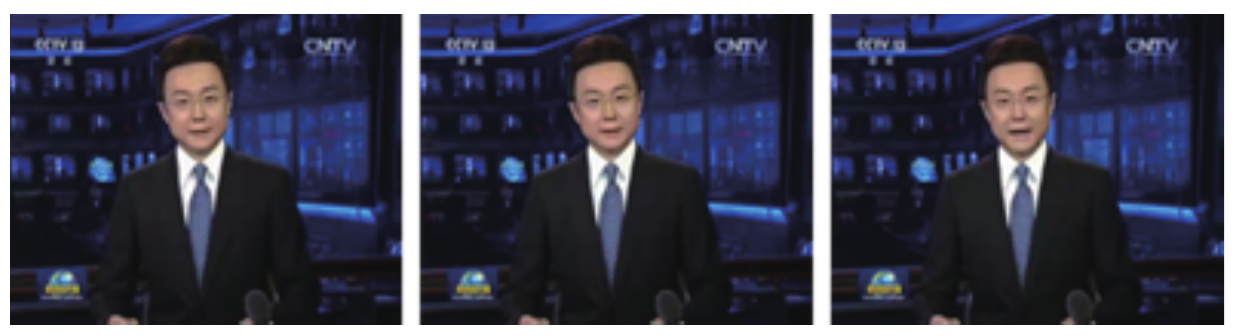}
\caption{ Static images from the CMLR dataset.}
\label{figure:data-CMLR}
\end{figure}

\begin{figure}[!t]
\centering
\includegraphics[width=3in]{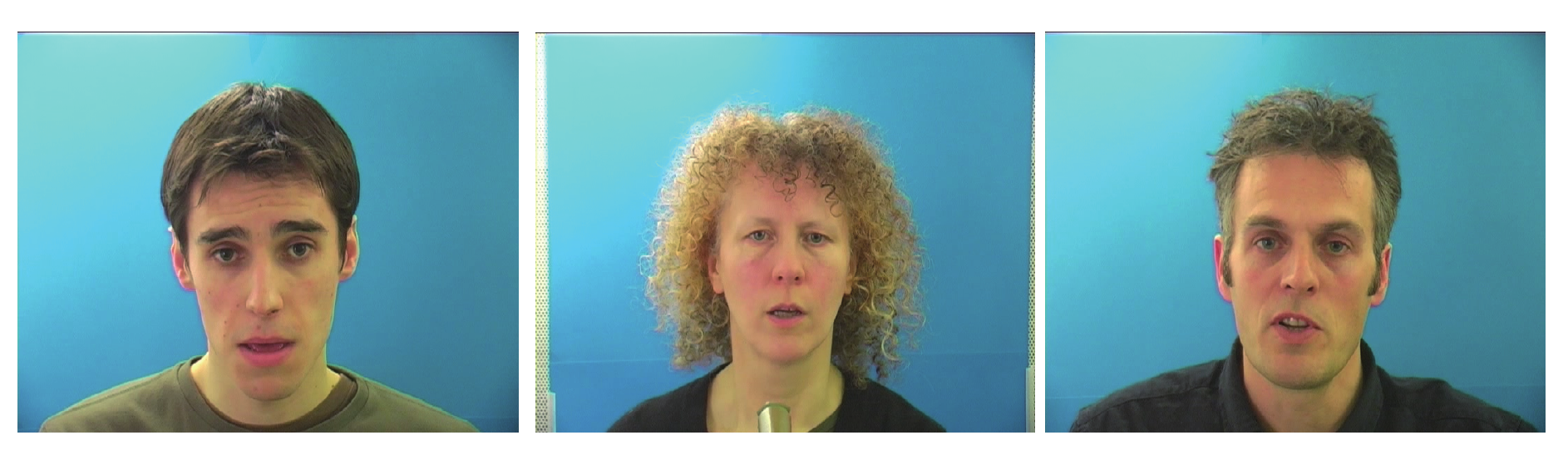}
\caption{ Static images from the GRID dataset.}
\label{figure:data-GRID}
\end{figure}

\subsection{Dataset}
The Chinese Mandarin Lip Reading (CMLR) dataset \cite{zhao2019cascade} is the largest Chinese Mandarin sentence lipreading dataset (example images are shown in Fig \ref{figure:data-CMLR}). The whole dataset contains sequences recorded by 11 speakers. We split the dataset to form the training and test set with 9 speakers and 2 speakers, respectively. Note the speakers are non-overlap in training and test. For experiments on overlap speakers, we divided the training, valid and test sets by following \cite{zhao2019cascade}. The CMLR division protocol is shown in Table \ref{tab:data-CMLR}.

\begin{table}[h]
\fontsize{10}{10}\selectfont
    \caption{Statistics for the CMLR dataset.}
    \label{tab:data-CMLR}
    \centering
    \begin{tabular}{llcc}
        \toprule
        &\textbf{Set} & \textbf{Speakers}& \textbf{Sentences}\\
        \midrule
        \multirow{2}*{Unseen}&Train & 9 & 81094    \\ 
        &Test & 2 & 20978  \\
        \midrule
        \multirow{2}*{Overlap}&Train & 11 & 71452   \\ 
        &Test & 11 & 20418   \\
        \bottomrule
    \end{tabular}
\end{table}

The GRID dataset \cite{cooke2006audio} has 33 speakers recorded, and is a widely used dataset in lipreading (example images are shown in Fig \ref{figure:data-GRID}). Each sentences consists of a sequence of verb + color + preposition + letter + number + adverb. e.g. "bin blue at f five again". To split the datasets into unseen and overlap speakers, we follow the setting as suggested in \cite{assael2016lipnet}. The division protocol are provided in Table \ref{tab:data-GRID}.

\begin{table}[h]
\fontsize{10}{10}\selectfont
    \caption{Statistics for the GRID dataset.}
    \label{tab:data-GRID}
    \centering
    \begin{tabular}{llcc}
        \toprule
        &\textbf{Set} & \textbf{Speakers}& \textbf{Sentences}\\
        \midrule
        \multirow{2}*{Unseen}&Train & 29 & 28837    \\ 
        &Test & 4 & 3986  \\
        \midrule
        \multirow{2}*{Overlap}&Train & 33 & 24408   \\ 
        &Test & 33 & 8415   \\
        \bottomrule
    \end{tabular}
\end{table}

\subsection{Evaluation Metrics}
To measure the performance of the proposed method and the baselines, we adopt the widely used evaluation metrics in Automatic Speech Recognition: Word Error Rate (WER) and Character Error Rate (CER). WER/CER is defined as the minimum number of word/character operations (including substitution, deletion, and insertion operations), which are required to convert the predicted label into the ground truth, and then divided by the number of words/characters in the ground truth. The calculation is defined as follows:
\begin{equation}
WER/CER=100\cdot \frac{S+D+I}{N},
\end{equation}
where $S$ denotes the substitution, $D$ represents the deletion, $I$ denotes the insertion, and $N$ is the number of words in the ground truth. Note that a smaller WER/CER indicates a higher prediction accuracy. In addition, the evaluation metric we use on the Chinese dataset CMLR is CER only, where CER means the Character Error Rate. The CER is calculated in the same way as WER, that is, each Chinese character is regarded as an English word.

\subsection{Implementation Details}
\begin{figure}[!t]
\centering
\includegraphics[width=3in]{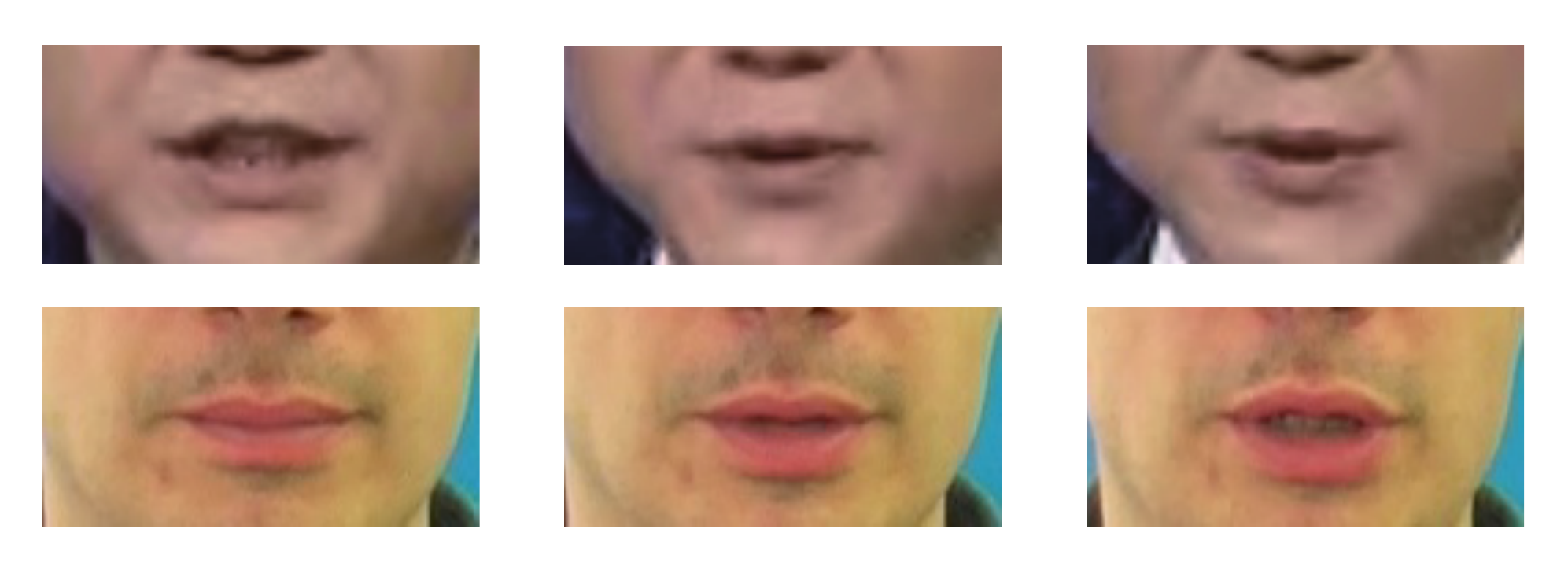}
\caption{ Examples of the mouth-centered crop.}
\label{figure:data-1}
\end{figure}

For each video clip, we first extracted the face image in each frame using the DLib face detector \cite{amodio2018automatic}. The coordinates of the landmarks generated by the face detector are used as the input of the landmark branch. The affine transformation is applied to each face image to obtain 160 × 80-pixel mouth-centered crop as the input to the visual branch. (see Fig \ref{figure:data-1}). This experiment uses the Adam optimizer to optimize the parameters with an initial learning rate of 0.0003. When each training error did not improve within 4 epochs, the learning rate decreases by 50\begin{math} \%  \end{math}.

\subsection{Competitors}

We compared our method with several state-of-the-art methods: LipNet\cite{assael2016lipnet}, CSSMCM\cite{zhao2019cascade}, CALLip\cite{huang2021callip}, LCSNet\cite{xue2022lcsnet} and WAS\cite{son2017lip}.

\noindent \textbf{LipNet}: LipNet is the first end-to-end sentence-level lipreading model that achieves an accuracy of 95.2 on the GRID dataset.

\noindent \textbf{CSSMCM}: This model is specifically designed for Chinese lipreading, combining factors such as pinyin and tones to help predict Chinese characters.

\noindent \textbf{CALLip}: CALLip improves the performance of the model by introducing attribute learning and contrast learning into the sentence-level lipreading pipeline, which using an attribute learning module to extract speaker identity features and eliminate cross-speaker variations. 

\noindent \textbf{LCSNet}: This model extract features that are more relevant to the lip motion by the channel attention module and the selective feature fusion module to improve recognition accuracy.

\noindent \textbf{WAS}: This baseline uses video information to predict sentences by seq2seq model, and achieves advanced performance on the LRS dataset.

\begin{table}[h]
\fontsize{10}{10}\selectfont
\caption{Performance comparison with state-of-the-arts on CMLR. -: results not available. Best results are in boldface.}
\label{tab:SOTA-CMLR}
    \centering
    \begin{tabular}{lcc}
        \toprule
      \textbf{Methods} & \textbf{Unseen}& \textbf{Overlap}\\
        \midrule
        LipNet &  52.18  & 33.41  \\ 
    WAS & - &38.93 \\
    CSSMCM & 50.08 & 32.48  \\
    CALLip & - &  31.18 \\
    LCSNet & 46.98 & 30.03 \\
    LipFormer&  \textbf{43.18} &  \textbf{27.79}\\
        \bottomrule
    \end{tabular}
\end{table}

\begin{table}[h]
\fontsize{10}{10}\selectfont
\caption{Performance comparison with state-of-the arts on GRID. Best results are in boldface.}
\label{tab:SOTA-GRID}
    \centering
    \begin{tabular}{lcc}
        \toprule
      \textbf{Methods} & \textbf{Unseen}& \textbf{Overlap}\\
        \midrule
        LipNet & 17.5 & 4.8   \\ 
     WAS & 14.6 & 3.0  \\
     CALLip & - & 2.48  \\
     LCSNet & 11.6 & 2.3 \\
     LipFormer &\textbf{9.64} & \textbf{1.45}  \\  
        \bottomrule
    \end{tabular}
\end{table}

\subsection{Comparison with Competitors}

To prove the effectiveness of the proposed method, we first compare it to SOTA competitors on the CMLR and GRID datasets. We empirically observed that the CTC loss could result in non-convergence during the training on the CMLR dataset. Hence, we replaced the CTC with the cascaded Seq2Seq module. We report the results for both the unseen and overlap speakers of two datasets. Experimental results are shown in Tables \ref{tab:SOTA-CMLR} and \ref{tab:SOTA-GRID}, respectively.

Table \ref{tab:SOTA-CMLR} shows that the LipFormer has a notable improvement over the state-of-the-art methods for \textit{unseen} speakers. Comparing to LipNet, LipFormer uses landmarks to jointly describe the lip motion. The comparison results show that the adoption of multi-modal features is beneficial to performance improvement. Compared with CSSMCM, the WER of LipFormer is further reduced by 6.9\begin{math} \%  \end{math}. One reason is that LipFormer learned more consistent features of lip motion with multi-modal features, so that the model is well-generalized to the unseen speakers. For the overlap speakers, LipFormer outperforms the SOTA methods on the CMLR dataset and achieves a character error rate of 27.79\begin{math} \%  \end{math}. It shows that LipFormer is well-suited to both unseen and overlap speakers.

Table \ref{tab:SOTA-GRID} shows the comparison results of different methods on the GRID dataset. Compared with LipNet, LipFormer can reduced the WER by 7.86\begin{math} \%  \end{math} for unseen speakers and WER by 3.35\begin{math} \%  \end{math} for overlap speakers. This affirms that landmarks can help solving the generalisation problem. It can also be observed that LipFormer outperforms other methods for both unseen and overlap speakers, even thought these speakers are from different ethnics.

\begin{table}[h]
\fontsize{10}{10}\selectfont
\caption{Performance of LipFormer and its variants on CMLR. The front-end structure of LipFormer is Visual-only+Landmark+Transformer. Best results are in boldface.}
\label{tab: freq}
    \centering
    \begin{tabular}{llcc}
        \toprule
     \textbf{\textbf{\#}} & \textbf{Methods} & \textbf{Unseen}& \textbf{Overlap}\\
        \midrule
       1 & Visual-only & 48.14 & 29.15   \\ 
     2 &Visual-only+Landmark & 43.48 & 28.0  \\
     3 &LipFormer &\textbf{43.18} & \textbf{27.79}  \\
        \bottomrule
    \end{tabular}
\end{table}

\begin{table}[h]
\fontsize{10}{10}\selectfont
\caption{Performance of LipFormer and its variants on GRID. Best results are in boldface.}
\label{tab: freq1}
    \centering
    \begin{tabular}{llcc}
        \toprule
      \textbf{\textbf{\#}} &\textbf{Methods} & \textbf{Unseen}& \textbf{Overlap}\\
        \midrule
        1 & Visual-only &  12.35& 2.82   \\ 
      2 &Visual-only+Landmark & 10.24 & 2.2  \\
      3 &LipFormer &\textbf{9.64} & \textbf{1.45}  \\
     \bottomrule
    \end{tabular}
\end{table}

\begin{figure}[!t]
\centering
\subfloat[]{\includegraphics[width=1.7in]{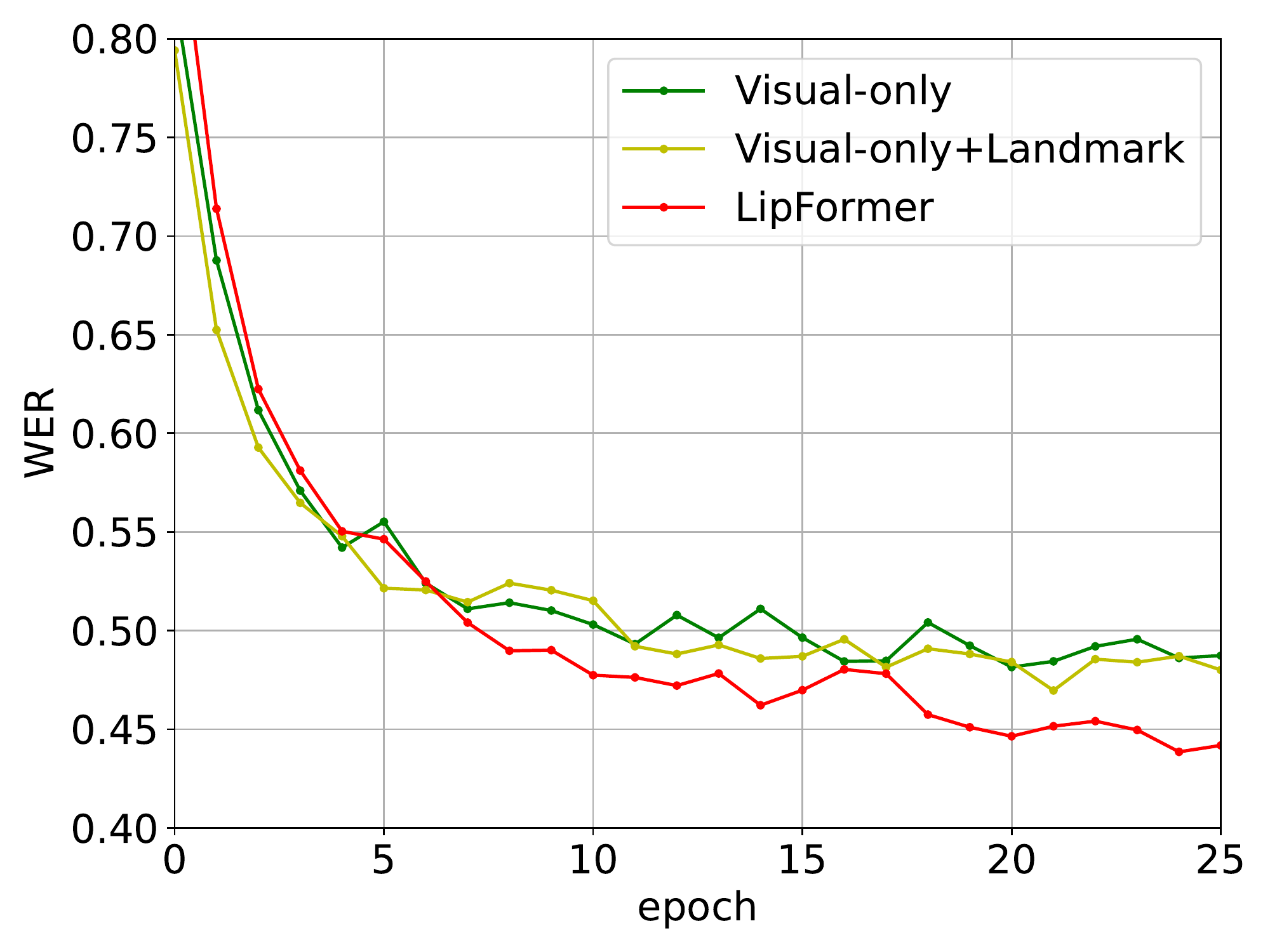}%
\label{fig_cer_CMLR}}
\hfil
\subfloat[]{\includegraphics[width=1.7in]{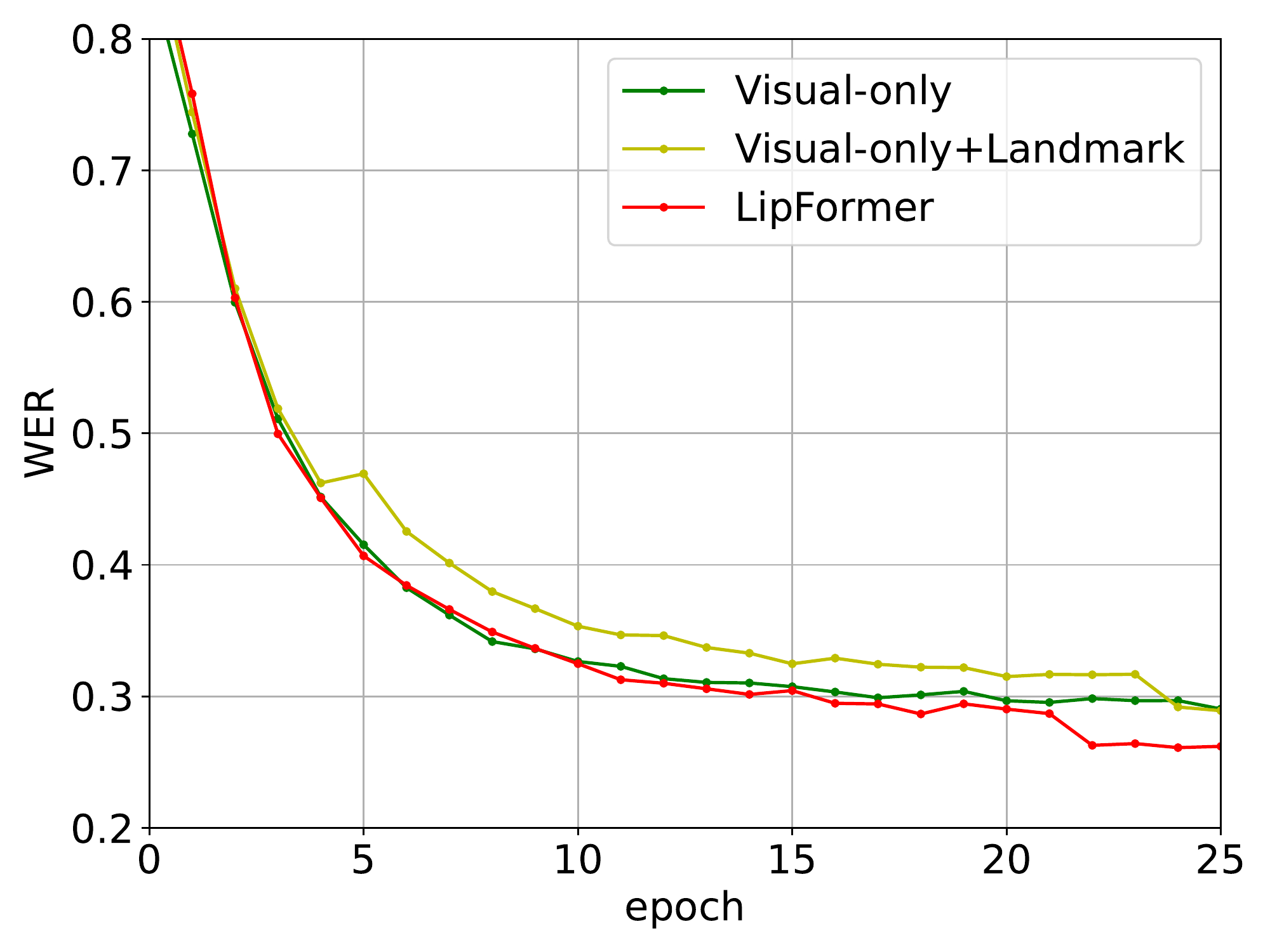}%
\label{fig_second_case}}
\caption{Performance of model variants on CMLR dataset. (a) Unseen. (b) Overlap.}
\label{fig_cer_CMLR}
\end{figure}

\begin{figure}[!t]
\centering
\subfloat[]{\includegraphics[width=1.7in]{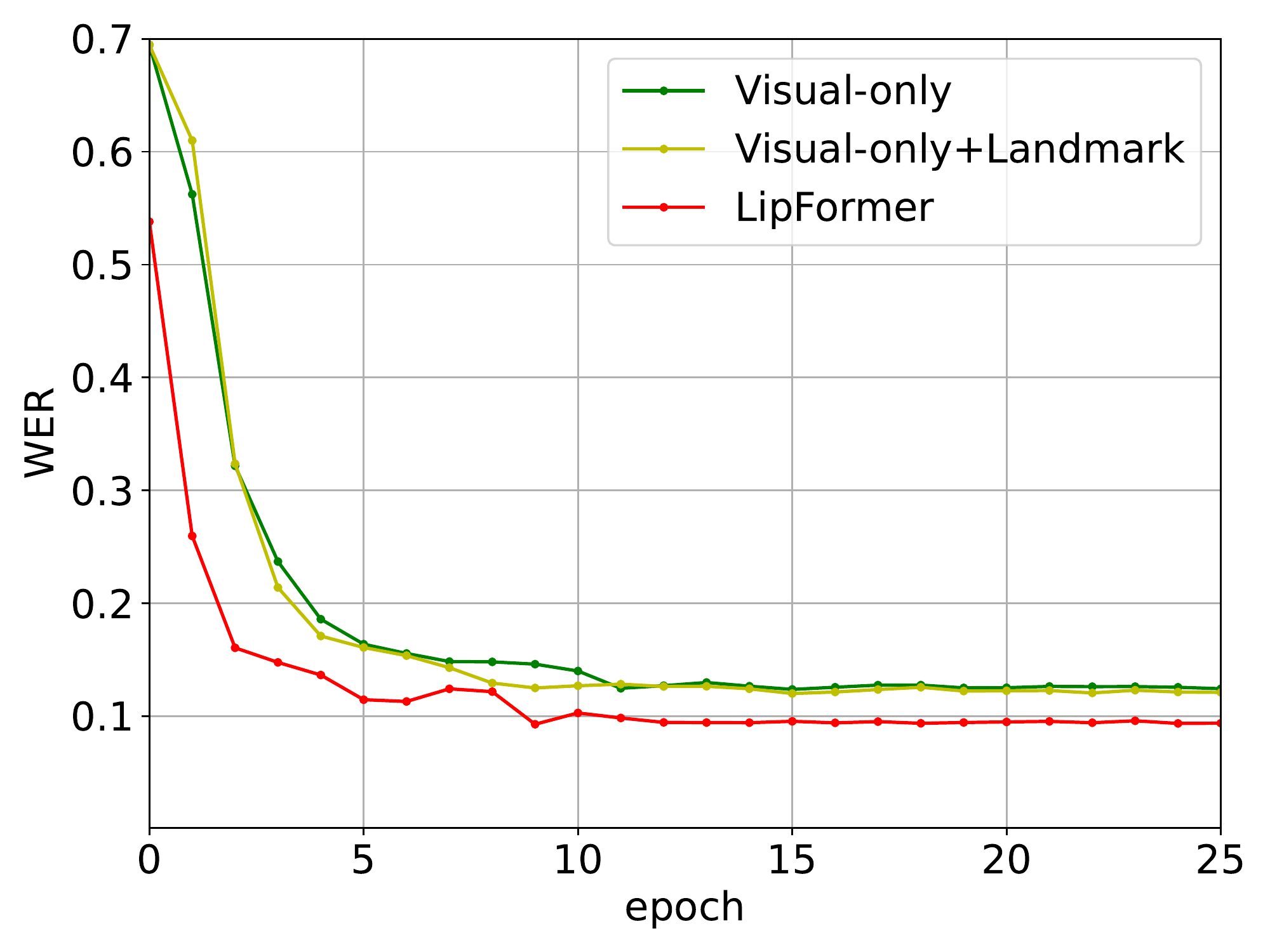}%
\label{fig_cer_GRID}}
\hfil
\subfloat[]{\includegraphics[width=1.7in]{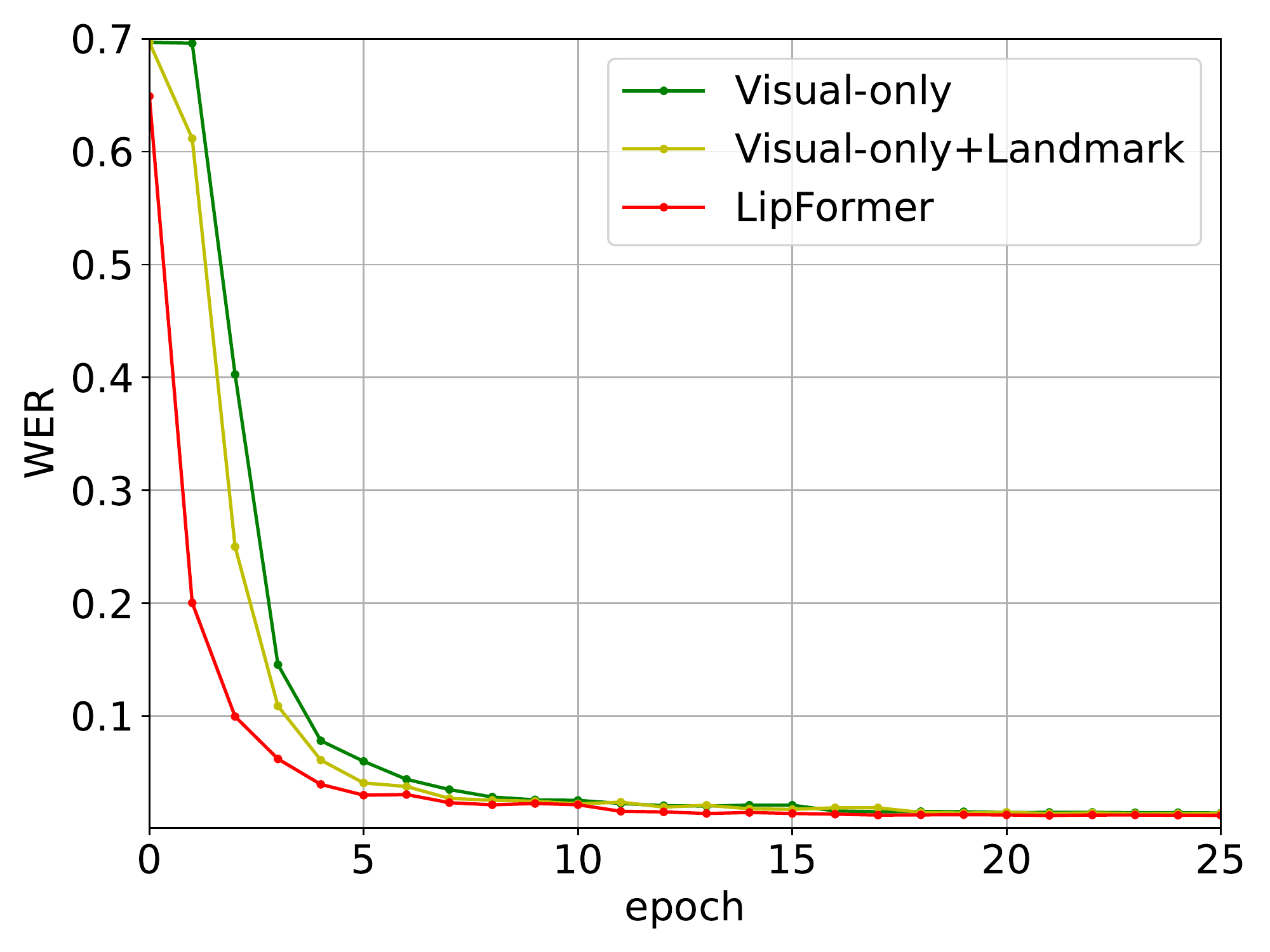}%
\label{fig_cer_GRID}}
\caption{Performance of model variants on GRID dataset. (a) Unseen. (b) Overlap.}
\label{fig_cer_GRID}
\end{figure}

\subsection{Ablation study}
To verify that the landmark branch and transformer module can effectively improve the lipreading accuracy for unseen speakers, we decouple the LipFormer framework and tested them on the CMLR and GRID datasets. The experimental results are shown in Tables \ref{tab: freq} and \ref{tab: freq1}, respectively. Fig \ref{fig_cer_CMLR} and Fig \ref{fig_cer_GRID} shows the WER curves of model variants on two datasets.

\begin{table}[h]
\fontsize{10}{10}\selectfont
\caption{Performance of models with different embedding sizes on CMLR dataset.}
\label{tab:hy}
    \centering
    \begin{tabular}{llcc}
        \toprule
     \textbf{\textbf{\#}} & \textbf{Methods} & \textbf{Unseen}& \textbf{Overlap}\\
        \midrule
       1 & 256 & 43.61 & 28.34   \\ 
     2 &512 &\textbf{43.18} & \textbf{27.79}  \\
     3 &1024 & 43.47 & 28.12  \\
        \bottomrule
    \end{tabular}
\end{table}

\begin{table*}[h]
\caption{Prediction of different models on CMLR and GRID.}
\label{tab:case}
    \centering
    \begin{tabular}{l|c|c|c|c}
        \hline
        \multirow{2}*{Methods}  & \multicolumn{2}{c|}{CMLR} & \multicolumn{2}{c}{GRID}\\
        \cline{2-5}
          &       Unseen  & Overlap & Unseen & Overlap \\
        \hline
        Ground truth &\begin{CJK}{UTF8}{gbsn} 人口较少名族发展规划\end{CJK} & \begin{CJK}{UTF8}{gbsn} 一些新做法取代了老传统\end{CJK}&bin blue at f three please  &  bin blue at d two soon    \\
        Visual-only&\begin{CJK}{UTF8}{gbsn} \textcolor{red}{中外}较少名族发展\textcolor{red}{模范}\end{CJK} & \begin{CJK}{UTF8}{gbsn} \textcolor{red}{议程改进}取代了\textcolor{red}{措施}\end{CJK}&bin blue at \textcolor{red}{s} three please  &  bin blue by d \textcolor{red}{five} soon    \\
     Visual-only+Landmark & \begin{CJK}{UTF8}{gbsn} 人口较少名族\textcolor{red}{检查}规划\end{CJK} & \begin{CJK}{UTF8}{gbsn} 一些新做法取代了\textcolor{red}{老烧}\end{CJK}  & bin blue at \textcolor{red}{l} three please & bin blue at d two \textcolor{red}{again}  \\
     LipFormer &\begin{CJK}{UTF8}{gbsn} 人口较少名族发展规划\end{CJK} & \begin{CJK}{UTF8}{gbsn} 一些新做法取代了老传统\end{CJK}& bin blue at f three please  &  bin blue at d two soon    \\ 
     \hline   
    \end{tabular}
   
\end{table*}
The variants of our method, i.e., \#1 and \#2, which fuse the visual branch and landmark branch outperform that of the model with visual-only branch. Specifically, compared with visual-only model, for the unseen and overlap speakers, the WER of method \#2 reduced by 4.66\begin{math} \%  \end{math} and 1.15\begin{math} \%  \end{math} on the CMLR dataset and reduced by 2.11\begin{math} \%  \end{math} and 0.62\begin{math} \%  \end{math} on the GRID dataset, respectively.  This demonstrates that landmark can normalizes the lip shapes of different speakers, eliminate the irrelevant visual variations, enhance the generalization ability of the model to unseen speakers. Comparing LipFormer  with Method \#2, experimental results show that the transformer module learns the correspondence between visual-landmark embedding to achieve fusion of cross-modal heterogeneous features, further improve the performance of the model. 

\subsection{Sensitivity to Hyper-parameter}
In this section, we conduct experiments on CMLR to investigate the effect of hyper-parameter in LipFormer on the model performance. The feature extractor encodes each time step as a feature embedding. To investigate the size of the feature embedding in the landmark branch affects the learning performance of the model, to this end, the size of embedding is controlled by controlling the number of output channels of the GRU in the landmark branch. Three different sizes are designed:256, 512, 1024. 


Table \ref{tab:hy} summarize the performance of each model for different embedding sizes. In general, as the number of embedding increases, the performance of the model will be better. However, a too-large embedding size may cause overfitting of the model and the performance of the model may decrease instead of increasing. The optimal number of channels for the first layer of GRU output is 512 on the CMLR dataset both for unseen speakers and overlap speakers.

\subsection{Case Study}
To qualitatively analyze the performance of the proposed model, this section evaluates a part of the predicted results. Table \ref{tab:case} show some sentences generated by each variant model on the CMLR and GRID datasets, where the characters highlighted in red are incorrect.

It can be seen from Table \ref{tab:case} that there are some differences between the sentences generated by Visual-only model and the ground truth on the CMLR dataset, such as predicting\begin{CJK}{UTF8}{gbsn}"人口"\end{CJK} (which means population) as \begin{CJK}{UTF8}{gbsn}"中外"\end{CJK}(which means Chinese-foreign), predicting \begin{CJK}{UTF8}{gbsn}"一些"\end{CJK} (which means some) as \begin{CJK}{UTF8}{gbsn}"议程"\end{CJK} (which means agenda) ,etc. This shows that the model translates different texts when people say the same word due to the visual variations such as lip shape. Visual-only+Landmark model generates sentences that are closer to the ground truth, which indicates that landmark can increase the accuracy of different speakers when pronouncing the same sentence. The sentences generated by LipFormer are correct, This shows that Transformer can promote the fusion of cross modal information. On the GRID dataset, some letters are easy to predict errors, for example, ”f” is predicted to be ”s”. These errors are mainly caused by letters with similar pronunciations.

\subsection{Alignment Visualization}

\begin{figure}[!t]
\centering
\subfloat[]{\includegraphics[width=1.7in]{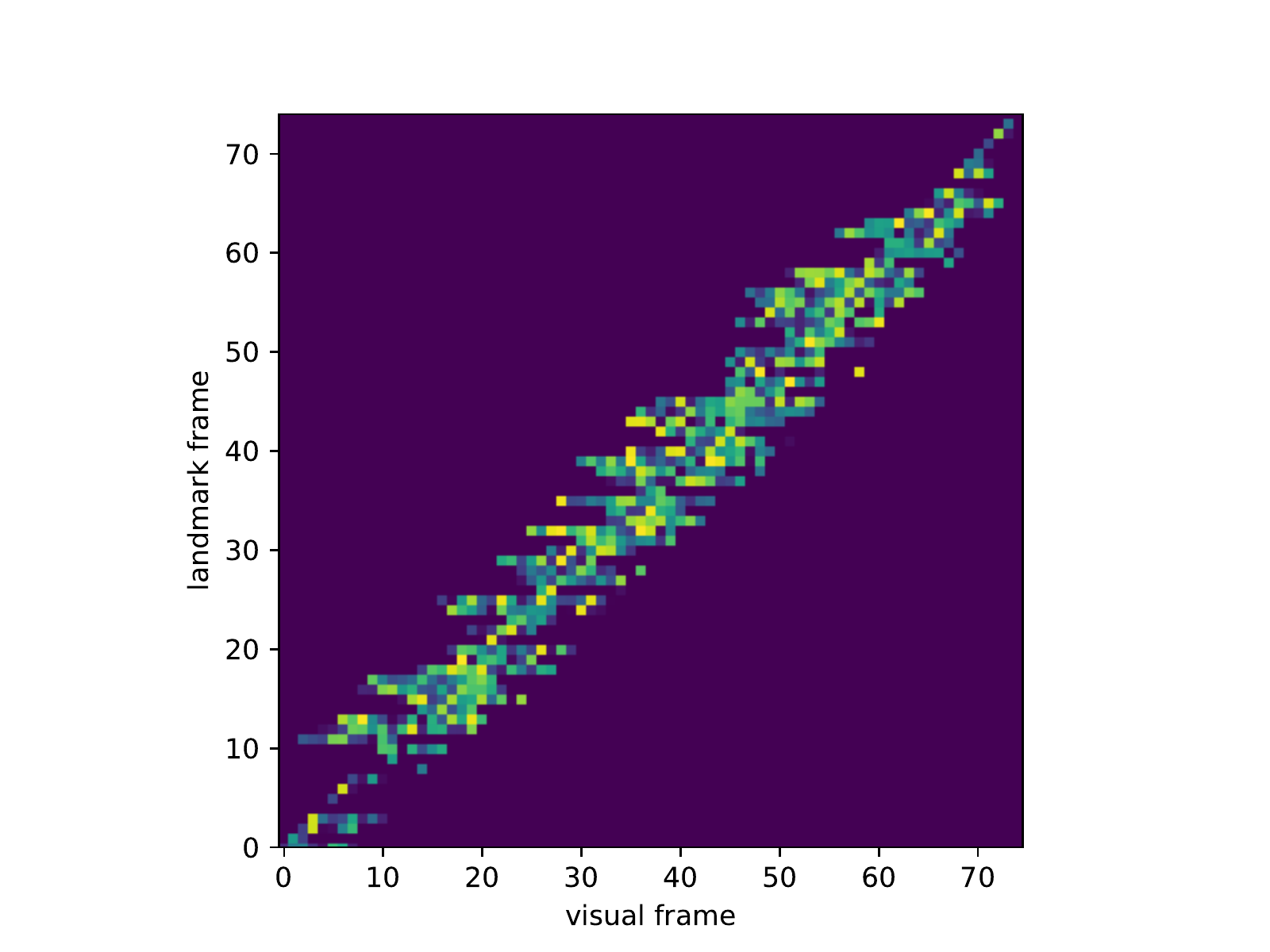}%
\label{fig_align_75}}
\hfil
\subfloat[]{\includegraphics[width=1.7in]{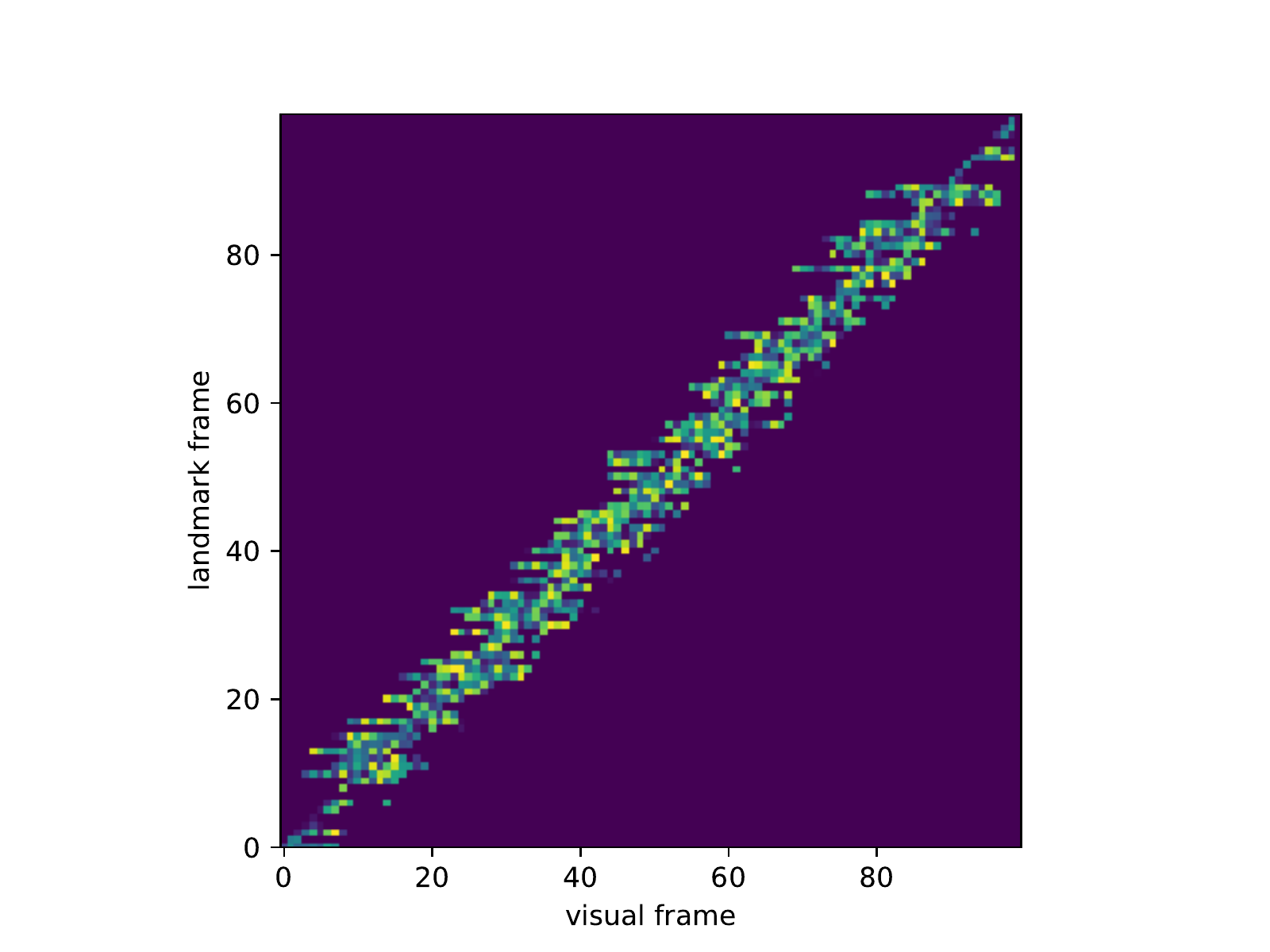}%
\label{fig_align_100}}
\caption{Illustration of the alignment of visual-landmark embedding on the CMLR dataset. Where the vertical axis represents the visual embedding and the horizontal axis represents the landmark embedding.}
\label{fig_align}
\end{figure}

Figure \ref{fig_align_75} and Figure \ref{fig_align_100} visualize the alignment between visuals and landmarks at 75 and 100 frames on the CMLR dataset by the cross-attention module, respectively. Each row in the figure represents a visual modality, and each column represents a landmark modality. The highlighted area in the figure indicates the degree of alignment between the visual-landmark feature embedding during feature fusion.

Figure \ref{fig_align} shows that in the process of feature fusion, cross-modal attention can learn the corresponding relationship between two modal feature embeddings, achieve the alignment between visual-landmark embedding, the diagonal trend is obvious. When pronouncing the same word, attention is concentrated on the corresponding different modal frames.

\section{Conclusion}\label{sec:con}
In this paper, we propose a cross-modal Transformer framework for sentence-level lipreading, which can generalize to unseen speakers by using landmarks as motion trajectories to calibre the visual variations. The model can level up the alignment of heterogeneous features by  cross-modal fusion suggested by the cross-attention. Extensive experimental results show that our framework can effectively generalize to unseen speakers. In future work, we put forward the relevance of this research in more challenging datasets, e.g., side-view speakers.



%

\bibliographystyle{IEEEtran}
\bibliography{sample}

\begin{thebibliography}{10}
\providecommand{\url}[1]{#1}
\csname url@samestyle\endcsname
\providecommand{\newblock}{\relax}
\providecommand{\bibinfo}[2]{#2}
\providecommand{\BIBentrySTDinterwordspacing}{\spaceskip=0pt\relax}
\providecommand{\BIBentryALTinterwordstretchfactor}{4}
\providecommand{\BIBentryALTinterwordspacing}{\spaceskip=\fontdimen2\font plus
\BIBentryALTinterwordstretchfactor\fontdimen3\font minus
  \fontdimen4\font\relax}
\providecommand{\BIBforeignlanguage}[2]{{%
\expandafter\ifx\csname l@#1\endcsname\relax
\typeout{** WARNING: IEEEtran.bst: No hyphenation pattern has been}%
\typeout{** loaded for the language `#1'. Using the pattern for}%
\typeout{** the default language instead.}%
\else
\language=\csname l@#1\endcsname
\fi
#2}}
\providecommand{\BIBdecl}{\relax}
\BIBdecl

\bibitem{2019Hearing}
Y.~Zhao, R.~Xu, X.~Wang, P.~Hou, H.~Tang, and M.~Song, ``Hearing lips:
  Improving lip reading by distilling speech recognizers,'' 2019.

\bibitem{zhang2019understanding}
X.~Zhang, H.~Gong, X.~Dai, F.~Yang, N.~Liu, and M.~Liu, ``Understanding
  pictograph with facial features: end-to-end sentence-level lip reading of
  chinese,'' in \emph{Proceedings of the AAAI Conference on Artificial
  Intelligence}, vol.~33, no.~01, 2019, pp. 9211--9218.

\bibitem{hilder2009comparison}
S.~Hilder, R.~W. Harvey, and B.-J. Theobald, ``Comparison of human and
  machine-based lip-reading.'' in \emph{AVSP}, 2009, pp. 86--89.

\bibitem{KIM2004295}
J.~O. Kim, W.~Lee, J.~Hwang, K.~S. Baik, and C.~H. Chung, ``Lip print
  recognition for security systems by multi-resolution architecture,''
  \emph{Future Generation Computer Systems}, vol.~20, no.~2, pp. 295--301,
  2004.

\bibitem{2004Lip}
O.~K. Jin, W.~Lee, J.~Hwang, K.~S. Baik, and C.~H. Chung, ``Lip print
  recognition for security systems by multi-resolution architecture,''
  \emph{Future Generation Computer Systems}, vol.~20, no.~2, pp. 295--301,
  2004.

\bibitem{2020Discriminative}
B.~Xu, C.~Lu, Y.~Guo, and J.~Wang, ``Discriminative multi-modality speech
  recognition,'' 2020.

\bibitem{petridis2018end}
S.~Petridis, T.~Stafylakis, P.~Ma, F.~Cai, G.~Tzimiropoulos, and M.~Pantic,
  ``End-to-end audiovisual speech recognition,'' in \emph{2018 IEEE
  international conference on acoustics, speech and signal processing
  (ICASSP)}.\hskip 1em plus 0.5em minus 0.4em\relax IEEE, 2018, pp. 6548--6552.

\bibitem{stafylakis2017combining}
T.~Stafylakis and G.~Tzimiropoulos, ``Combining residual networks with lstms
  for lipreading,'' \emph{arXiv preprint arXiv:1703.04105}, 2017.

\bibitem{zhang2021efficient}
T.~Zhang, L.~He, X.~Li, and G.~Feng, ``Efficient end-to-end sentence-level
  lipreading with temporal convolutional networks,'' \emph{Applied Sciences},
  vol.~11, no.~15, p. 6975, 2021.

\bibitem{zhao2020mutual}
X.~Zhao, S.~Yang, S.~Shan, and X.~Chen, ``Mutual information maximization for
  effective lip reading,'' in \emph{2020 15th IEEE International Conference on
  Automatic Face and Gesture Recognition (FG 2020)}.\hskip 1em plus 0.5em minus
  0.4em\relax IEEE, 2020, pp. 420--427.

\bibitem{zhao2019cascade}
Y.~Zhao, R.~Xu, and M.~Song, ``A cascade sequence-to-sequence model for chinese
  mandarin lip reading,'' in \emph{Proceedings of the ACM Multimedia Asia},
  2019, pp. 1--6.

\bibitem{xu2018lcanet}
K.~Xu, D.~Li, N.~Cassimatis, and X.~Wang, ``Lcanet: End-to-end lipreading with
  cascaded attention-ctc,'' in \emph{2018 13th IEEE International Conference on
  Automatic Face \& Gesture Recognition (FG 2018)}.\hskip 1em plus 0.5em minus
  0.4em\relax IEEE, 2018, pp. 548--555.

\bibitem{liu2020fastlr}
J.~Liu, Y.~Ren, Z.~Zhao, C.~Zhang, B.~Huai, and J.~Yuan, ``Fastlr:
  Non-autoregressive lipreading model with integrate-and-fire,'' in
  \emph{Proceedings of the 28th ACM International Conference on Multimedia},
  2020, pp. 4328--4336.

\bibitem{assael2016lipnet}
Y.~M. Assael, B.~Shillingford, S.~Whiteson, and N.~De~Freitas, ``Lipnet:
  End-to-end sentence-level lipreading,'' \emph{arXiv preprint
  arXiv:1611.01599}, 2016.

\bibitem{2014Return}
K.~Chatfield, K.~Simonyan, A.~Vedaldi, and A.~Zisserman, ``Return of the devil
  in the details: Delving deep into convolutional nets,'' \emph{arXiv
  e-prints}, 2014.

\bibitem{2014Empirical}
J.~Chung, C.~Gulcehre, K.~H. Cho, and Y.~Bengio, ``Empirical evaluation of
  gated recurrent neural networks on sequence modeling,'' \emph{Eprint Arxiv},
  2014.

\bibitem{0Connectionist}
A.~Auvolat and T.~Mesnard, ``Connectionist temporal classification: Labelling
  unsegmented sequences with recurrent neural networks,'' 2006.

\bibitem{cooke2006audio}
M.~Cooke, J.~Barker, S.~Cunningham, and X.~Shao, ``An audio-visual corpus for
  speech perception and automatic speech recognition,'' \emph{The Journal of
  the Acoustical Society of America}, vol. 120, no.~5, pp. 2421--2424, 2006.

\bibitem{huang2021callip}
Y.~Huang, X.~Liang, and C.~Fang, ``Callip: Lipreading using contrastive and
  attribute learning,'' in \emph{Proceedings of the 29th ACM International
  Conference on Multimedia}, 2021, pp. 2492--2500.

\bibitem{zhang2018unsupervised}
Y.~Zhang, Y.~Guo, Y.~Jin, Y.~Luo, Z.~He, and H.~Lee, ``Unsupervised discovery
  of object landmarks as structural representations,'' in \emph{Proceedings of
  the IEEE Conference on Computer Vision and Pattern Recognition}, 2018, pp.
  2694--2703.

\bibitem{haghpanah2022real}
M.~A. Haghpanah, E.~Saeedizade, M.~T. Masouleh, and A.~Kalhor, ``Real-time
  facial expression recognition using facial landmarks and neural networks,''
  in \emph{2022 International Conference on Machine Vision and Image Processing
  (MVIP)}.\hskip 1em plus 0.5em minus 0.4em\relax IEEE, 2022, pp. 1--7.

\bibitem{2012On}
V.~Estellers, M.~Gurban, and J.~P. Thiran, ``On dynamic stream weighting for
  audio-visual speech recognition,'' \emph{IEEE Transactions on Audio Speech
  and Language Processing}, vol.~20, no.~4, pp. 1145--1157, 2012.

\bibitem{2008Patch}
P.~J. Lucey, G.~Potamianos, and S.~Sridharan, ``Patch-based analysis of visual
  speech from multiple views,'' in \emph{Proceedings of the International
  Conference on Auditory-Visual Speech Processing 2008}, 2008.

\bibitem{sheerman2011cultural}
T.~Sheerman-Chase, E.-J. Ong, and R.~Bowden, ``Cultural factors in the
  regression of non-verbal communication perception,'' in \emph{2011 IEEE
  international conference on computer vision workshops (ICCV
  workshops)}.\hskip 1em plus 0.5em minus 0.4em\relax IEEE, 2011, pp.
  1242--1249.

\bibitem{1992Inferring}
G.~Papcun, J.~Hochberg, T.~R. Thomas, F.~Laroche, and S.~Levy, ``Inferring
  articulation and recognizing gestures from acoustics with a neural network
  trained on x‐ray microbeam data,'' \emph{Journal of the Acoustical Society
  of America}, vol.~92, no. 2 Pt 1, pp. 688--700, 1992.

\bibitem{2001Hmm}
M.~T. Chan, ``Hmm-based audio-visual speech recognition integrating geometric-
  and appearance-based visual features,'' in \emph{IEEE Fourth Workshop on
  Multimedia Signal Processing}, 2001.

\bibitem{1997Speechreading}
J.~Luettin and N.~A. Thacker, ``Speechreading using probabilistic models,''
  \emph{Computer Vision and Image Understanding}, vol.~65, no.~2, pp. 163--178,
  1997.

\bibitem{10.1007/978-3-319-54184-6_6}
J.~S. Chung and A.~Zisserman, ``Lip reading in the wild,'' in \emph{Computer
  Vision -- ACCV 2016}, 2017, pp. 87--103.

\bibitem{2018END}
I.~Fung and B.~Mak, ``End-to-end low-resource lip-reading with maxout cnn and
  lstm,'' 2018, pp. 2511--2515.

\bibitem{2017Improving}
M.~Wand and J.~Schmidhuber, ``Improving speaker-independent lipreading with
  domain-adversarial training,'' 2017.

\bibitem{2018Investigations}
M.~Wand, J.~Schmidhuber, and N.~T. Vu, ``Investigations on end- to-end
  audiovisual fusion,'' 2018.

\bibitem{son2017lip}
J.~Son~Chung, A.~Senior, O.~Vinyals, and A.~Zisserman, ``Lip reading sentences
  in the wild,'' in \emph{Proceedings of the IEEE conference on computer vision
  and pattern recognition}, 2017, pp. 6447--6456.

\bibitem{zhou2018syllable}
S.~Zhou, L.~Dong, S.~Xu, and B.~Xu, ``Syllable-based sequence-to-sequence
  speech recognition with the transformer in mandarin chinese,'' \emph{arXiv
  preprint arXiv:1804.10752}, 2018.

\bibitem{2020A}
S.~Ma, S.~Wang, and X.~Lin, ``A transformer-based model for sentence-level
  chinese mandarin lipreading,'' in \emph{2020 IEEE Fifth International
  Conference on Data Science in Cyberspace (DSC)}, 2020.

\bibitem{2022Visual}
P.~Ma, S.~Petridis, and M.~Pantic, ``Visual speech recognition for multiple
  languages in the wild,'' 2022.

\bibitem{3D-PersonVLAD}
L.~Wu, Y.~Wang, L.~Shao, and M.~Wang, ``3d personvlad: Learning deep global
  representations for video-based person re-identification,'' \emph{IEEE
  Transactions on Neural Networks and Learning Systems}, vol.~30, no.~11, pp.
  3347--3359, 2019.

\bibitem{WU-Co-attention}
L.~Wu, Y.~Wang, J.~Gao, M.~Wang, Z.-J. Zha, and D.~Tao, ``Deep co-attention
  based comparators for relative representation learning in person
  re-identification,'' \emph{IEEE Transactions on Neural Networks and Learning
  Systems}, vol.~32, no.~2, pp. 722--735, 2021.

\bibitem{Chen-MM-2022}
D.~Chen, M.~Wang, H.~Chen, L.~Wu, J.~Qin, and W.~Peng, ``Cross-modal retrieval
  with heterogeneous graph embedding,'' in \emph{Proceedings of the 30th ACM
  International Conference on Multimedia}, 2022, pp. 3291--3300.

\bibitem{amodio2018automatic}
A.~Amodio, M.~Ermidoro, D.~Maggi, S.~Formentin, and S.~M. Savaresi, ``Automatic
  detection of driver impairment based on pupillary light reflex,'' \emph{IEEE
  transactions on intelligent transportation systems}, vol.~20, no.~8, pp.
  3038--3048, 2018.

\bibitem{xue2022lcsnet}
F.~xue, T.~Yang, K.~Liu, Z.~Hong, M.~Cao, D.~Guo, and R.~Hong, ``Lcsnet:
  End-to-end lipreading with channel-aware feature selection,'' \emph{ACM
  Transactions on Multimedia Computing, Communications, and Applications
  (TOMM)}, 2022.

\end{thebibliography}

\vfill

\end{document}